\documentclass{article}
\usepackage{ijcai17}
\usepackage{times}
\usepackage{helvet}
\usepackage{courier}
\usepackage{subfig}
\usepackage{latexsym,amssymb,amsmath,amsthm}
\usepackage{algorithm}
\usepackage[noend]{algpseudocode}
\usepackage{xspace}
\usepackage{url}
\usepackage{enumerate}
\usepackage[defblank]{paralist}
\usepackage{graphicx}
\graphicspath{{imgs/}}
\usepackage{microtype}
\usepackage{comment}
\usepackage{todonotes}
\usepackage{hyperref}
\usepackage{tabularx}
\usepackage{thesisSymbols}
\usepackage{booktabs}
\usepackage{multirow}
\usepackage{MnSymbol}

\newcommand{\myi}{(\emph{i})\xspace}
\newcommand{\myii}{(\emph{ii})\xspace}

\newcommand{\A}{\mathcal{A}}

 \renewcommand{\L}{\mathcal{L}}
\newcommand{\M}{\mathcal{M}} 
 \renewcommand{\P}{\mathcal{P}}
 
\renewcommand{\S}{\mathcal{S}} 
\newcommand{\U}{\mathcal{U}}

\newcommand{\limp}{\mathbin{\rightarrow}}

\newcommand{\Next}{\raisebox{-0.27ex}{\LARGE$\circ$}}

\renewcommand{\Until}{\mathop{\U}}
\newcommand{\Since}{\mathop{\S}}

\newcommand{\true}{\mathit{true}}

\newcommand{\false}{\mathit{false}}
\newcommand{\ttrue}{{\mathit{tt}}}
\newcommand{\ffalse}{\mathit{ff}}
\newcommand{\Last}{\mathit{last}}
\newcommand{\Ended}{\mathit{end}}
\newcommand{\length}{\mathit{length}}

\newcommand{\nnf}{\mathit{nnf}}

\newcommand{\BOX}[1]{ [#1]}
\newcommand{\DIAM}[1]{\langle #1 \rangle}

\newcommand{\prev}{{\circleddash}}
\newcommand{\gpast}{{\boxminus}}
\newcommand{\past}{{\diamondminus}}

\newcommand{\LTL}{{\sc ltl}\xspace}
\newcommand{\LTLf}{{\sc ltl}$_f$\xspace}
\newcommand{\PLTL}{{\sc pltl}\xspace}
\newcommand{\DFLTL}{$\$${\sc fltl}\xspace}

\newcommand{\LDLf}{{\sc ldl}$_f$\xspace}

\newcommand{\REGEX}{{\sc re}\xspace}
\newcommand{\PDL}{{\sc pdl}\xspace}

\newcommand{\FO}{{\sc fo}\xspace}
\newcommand{\MSO}{{\sc mso}\xspace}
\newcommand{\AFW}{{\sc afw}\xspace}
\newcommand{\NFA}{{\sc nfa}\xspace}
\newcommand{\DFA}{{\sc dfa}\xspace}

\newcommand{\Nat}{{\rm I\kern-.23em N}}
\newcommand{\Prop}{\P}

\newcommand{\Stop}{\mathit{stop}}

\newcommand{\Tr}{\mathit{Tr}}

\newcommand{\expand}{\textbf{\textit{E}}}
\newcommand{\ttt}{{\textbf{\textit{T}}}}
\newcommand{\fff}{{\textbf{\textit{\texttt{F}}}}}

\newcommand{\atomize}[1]{\texttt{"}\ensuremath{#1}\texttt{"}}

\newcommand{\algoname}{\textsc{{\LDLf}2\NFA}}

\newtheorem{theorem}{Theorem}

\newtheorem{definition}{Definition}

\def\qed{\hfill{\qedboxempty}} 
\def\qedboxempty{\vbox{\hrule\hbox{\vrule\kern3pt
\vbox{\kern3pt\kern3pt}\kern3pt\vrule}\hrule}}

\begin{document}
%
\title{Specifying Non-Markovian Rewards in MDPs Using LDL on Finite Traces (Preliminary Version)}
\author{\#3008}
 \author{
 Ronen Brafman\\
 Ben-Gurion University, 
 Beer-Sheva, 
 Israel\\
  \url{brafman@cs.bgu.ac.il}\\
 \And
  Giuseppe De Giacomo \&  Fabio Patrizi\\
  Sapienza Universit\`a di Roma, Italy\\
  \url{{degiacomo,patrizi}@dis.uniroma1.it}
  }
\maketitle

\begin{abstract}
  In Markov Decision Processes (MDPs), the reward obtained in a state
  depends on the properties of the last state and action.  This state
  dependency makes it difficult to reward more interesting long-term
  behaviors, such as always closing a door after it has been opened,
  or providing coffee only following a request.  Extending MDPs to
  handle such non-Markovian reward function was the subject of two
  previous lines of work, both using variants of LTL to specify the
  reward function and then compiling the new model back into a
  Markovian model.  Building upon recent progress in the theories of
  temporal logics over finite traces, we adopt \LDLf for specifying
  non-Markovian rewards and provide an elegant automata construction
  for building a Markovian model, which extends that of previous work
  and offers strong minimality and compositionality guarantees.
\end{abstract}

\section{Introduction}\label{sec:introduction}

Markov Decision Processes (MDPs) are a central model for
sequential decision making under uncertainty. They are used to model
and solve many real-world problems, and to address the problem of learning  to behave well in unknown environments.
The Markov assumption is a key element of this model.
It states that the effects of an action depend only
on the state in which it was executed, and that reward given at a
state depends only on the previous action and 
state.
It has long been observed~\cite{BBG96,ThiebauxGSPK06} that
many performance criteria call for more sophisticated reward functions that do not
depend on the last state only.
For example, we may want to reward a robot that \emph{eventually
delivers coffee each time it gets a  request;} or,
to reward the robot for keeping its operating environment tidy, e.g., by
\emph{closing doors after opening}; or, to ensure it will
\emph{access restricted areas only after having acquired the right
  permission}. All these rewards are non-Markovian.
Recently, Littman, in his IJCAI 2015 invited talk ``Programming Agents via
Rewards,'' advocated that it may actually be more convenient, from
a design 
perspective, to assign rewards to the satisfaction of
declarative temporal properties, 
rather than 
to states.


To extend MDPs with non-Markovian rewards we need a language for specifying such rewards. Markovian rewards are specified as a function $R$ from the previous state and action to the reals. $R$ can be specified using an explicit reward matrix, or implicitly, by associating a reward with properties of the last state and action. 
With non-Markovian rewards, an explicit representation is no longer possible, as the number of possible histories or futures of a state is infinite. Hence, we must use an implicit specification  that can express properties of past (or future) states. 
To date, two specification languages have been proposed. \cite{BBG96} suggests using a temporal logic of the past.
Whether state $s_i$ satisfies such a past temporal formula depends on the entire sequence of states leading up to it: $s_1,s_2,\ldots, s_{i}$. 
Thus, we can reward appropriate response to a ``bring-coffee" command
by associating a reward with the property {\em the ``bring-coffee" command was issued in the past, and now I have coffee\/}.
A second proposal, by \cite{ThiebauxGSPK06}, uses a temporal logic of the future with a special symbol to denote awarding a reward. 
At each step, one checks whether this symbol must be true in the current state, for the reward formula to be satisfied in the initial state.
If that is the case, the current state is rewarded. This language is a little less intuitive, and its semantics is more complicated. 

Existing MDP solution methods, possibly with the
exception of Monte-Carlo tree search algorithms, rely heavily on the
Markov assumption, and cannot be applied directly with non-Markovian rewards. 
To address this, both proposals above transform the non-Markovian model to an
equivalent Markovian one that can be solved using existing algorithms, 
by enriching the state with information about the past. For example, suppose we extend our state to record
whether a "bring-coffee" command was issued earlier. A
reward for bringing coffee in states that indicate that the
"bring-coffee" command was issued in the past, is now Markovian.  It
rewards the same behaviors as the earlier non-Markovian reward on the
original state.  We call a model obtained by extending the state
space of the original non-Markovian MDP, an {\em extended MDP}.

Using an extended MDP is a well-known idea.  
Since state space size 
affects the practical and theoretical complexity 
of most MDP solution algorithms, the main question is how to \emph{minimally} enrich the state
so as to make rewards Markovian. 
\cite{BBG96} provide algorithms for constructing an extended MDP that attempt to minimize size by reducing the amount of information about the past that is maintained. While their construction does not generate the minimal extended MDP, they allude to using automata minimization techniques to accomplish this. \cite{ThiebauxGSPK06}, instead, use a construction that  works well with forward search based algorithms, such as LAO*~\cite{LAO*} and LRTDP~\cite{LRTDP}. Unlike classical dynamic programming methods that require the entire state space a-priori, 
these algorithms generate reachable states only. 
With a good heuristic function, they often generate only a fraction of the state space. 
So, while the augmented search space they obtain is not minimal, 
because states are constructed on the fly during forward
search, their approach does not require a-priori enumeration of the state space, and never generates an unreachable state. 
They call this property {\em blind minimality\/}.

The aim of this paper is to bring to bear developments in the theory of temporal logic over finite traces to the problem of specifying and solving MDP with non-Markovian rewards. 
%
%
We adopt \LDLf, a temporal logic of the future interpreted over finite traces, which  extends \LTLf, the classical linear-time temporal logic over finite traces \cite{DegVa13}.  \LDLf has the same computational features of \LTLf but it is more expressive, as it captures monadic second order logic (\MSO) on finite traces (i.e., inductively defined properties), instead of first-order logic (\FO), as \LTLf. A number of techniques based on automata manipulation have been developed for \LDLf,
to address tasks such as satisfiability, model checking, reactive synthesis, and planning under full/partial observability \cite{DegVa13,DegVa15,DegVa16,TorresB15,CTMBM17}. We exploit such techniques to generate
an extended MDP with good properties.

Our formalism has three important advantages:
1.  {\bf Enhanced expressive power.} 
We move from linear-time temporal logics used by past authors to
\LDLf, paying no additional (worst-case)
complexity costs.  \LDLf can encode in polynomial time \LTLf, regular expressions (\REGEX), the past \LTL (\PLTL) of \cite{BBG96},  and virtually all examples of \cite{ThiebauxGSPK06}. Often, \LDLf can represent more compactly and more intuitively conditions specified in \LTLf\  or \PLTL.
Future logics are more commonly used in the model checking community, as they are considered more natural for expressing desirable properties. 
This is especially true with complex properties that require the power of \LDLf.
%
2. {\bf Minimality and Compositionality.} 
We generate a minimal equivalent extended MDP, 
exploiting existing techniques for constructing automata that track
the satisfiability of an \LDLf formula. This construction is relatively simple and
compositional: if a new reward formula is to be added, we
only need to optimize the corresponding automaton and add it to the
current (extended) MDP. If the current MDP was minimal, the
resulting (extended) MDP is minimal, too.
3. {\bf Blind Minimality.} The automaton used to identify when a
reward should be given can be constructed in a forward manner using
progression, providing blind minimality as in~\cite{ThiebauxGSPK06}.
If, instead, we want pure minimality, unlike the construction
of~\cite{BBG96}, we can exploit the forward construction to never generate unreachable states, 
before applying automata minimization. Hence, we have the best of both worlds.

\section{Background}

\noindent {\bf MDPs.\ } A Markov Decision Process (MDP) $\M = \langle S,A,Tr,R\rangle$
consists of a set $S$ of states, a set $A$ of actions, a transition
function $Tr:S\times A\rightarrow Prob(S)$ that returns for every state
$s$ and action $a$ a distribution over the next state. We can further
restrict actions to be defined on a subset of $S$ only, and use
$A(s)$ to denote the actions applicable in $s$.
The reward function,
$R:S\times A \rightarrow \mathbb{R}$, 
specifies the real-valued reward received by the agent when applying
action $a$ in state $s$. In this paper, states in $S$ are truth assignments 
to a set $\Prop$ of primitive propositions. Hence, if $\varphi$ is a propositional formula and $s$
a state, we can check whether $s\models\varphi$.

A solution to an MDP, called a {\em policy}, assigns an action to each
state, possibly conditioned on past states and actions. 
The {\em value\/} of policy $\rho$ at $s$, $v^{\rho}(s)$, is the expected sum of
(possibly discounted) rewards when starting at state $s$ and selecting
actions based on $\rho$. Every MDP has an {\em optimal} policy, $\rho^*$, 
i.e., one that maximizes the expected sum
of rewards for every starting state $s\in S$. In the
case of infinite horizon, there exists an optimal policy that is stationary $\rho: S \rightarrow A$
(i.e., $\rho$ depends only on the current state) and
deterministic~\cite{Puterman}. There are diverse methods for computing an
optimal policy. With the exception of online simulation-based methods,
they rely 
on the fact that both transitions and rewards depend
on the last state only -- i.e., are independent of earlier states and
actions ({\em Markov property}).  The theoretical and practical complexity of solution algorithms is
strongly impacted by $|S|$.


\vspace{1ex}
\noindent {\bf LTL$_f$ and LDL$_f$.\ } 
\LTLf is essentially 
\LTL \cite{Pnueli77} interpreted over finite, instead of infinite, traces.
\LTLf is as expressive as \FO
over finite traces
and star-free \REGEX, thus strictly less expressive than
\REGEX, which in turn are as expressive as \MSO 
over finite traces.  \REGEX themselves
are not convenient for expressing temporal specifications,
since, e.g., they miss direct constructs for negation and
conjunction. 
For this reason, \cite{DegVa13} introduced \LDLf (\emph{linear dynamic logic on finite traces}),
which merges \LTLf with \REGEX, through the syntax of the
well-known logic of programs \PDL, \emph{propositional dynamic logic}
\cite{FiLa79,HaKT00,Var11}, but interpreted over finite traces.

We consider a variant of \LDLf that works also on empty traces. 
Formally, \LDLf formulas $\varphi$ are built as follows:
\[\begin{array}{lcl}
\varphi &::=& \ttrue  \mid \lnot \varphi \mid \varphi_1 \land \varphi_2 \mid \DIAM{\rho}\varphi \\
\rho &::=& \phi \mid \varphi? \mid  \rho_1 + \rho_2 \mid \rho_1; \rho_2 \mid \rho^*
\end{array}
\]
where $\ttrue$ stands for logical true; $\phi$ is a propositional
formula over $\Prop$ (including $\true$, not to be confused with
$\ttrue$); $\rho$ denotes path expressions, which are \REGEX over
propositional formulas $\phi$ with the addition of the test construct
$\varphi?$ typical of \PDL.  We use abbreviations
$\BOX{\rho}\varphi\doteq\lnot\DIAM{\rho}{\lnot\varphi}$, as in \PDL,
$\ffalse\doteq \lnot\ttrue$, to denote false, and
$\phi \doteq \DIAM{\phi}\ttrue$ to denote the occurence of proposition
$\phi$.

Intuitively, $\DIAM{\rho}\varphi$ states that, from the current step
in the trace, there exists an execution satisfying the \REGEX $\rho$ 
such that its last step satisfies $\varphi$, while
$\BOX{\rho}\varphi$ states that, from the current step, all executions
satisfying the \REGEX $\rho$ are such that their last step
satisfies $\varphi$.
Tests are used to insert into the execution path checks for
satisfaction of additional \LDLf formulas.

The semantics of \LDLf is given in terms of \emph{finite
  traces}, i.e., finite sequences $\pi=\pi_0,\ldots,\pi_n$ of elements from the alphabet $2^\Prop$.
We define $\pi(i)\doteq \pi_i$, $\length(\pi)\doteq n+1$, and
$\pi(i,j)\doteq \pi_i,\pi_{i+1},\ldots,\pi_j$. 
When $j>n$, $\pi(i,j)\doteq\pi(i,n)$.

In decision processes, traces are usually sequences of states {\em and} actions, i.e.,
they have the form: $\langle s_0,a_1,s_1,\ldots,s_{n-1}, a_n\rangle $. 
These can still be represented as traces of the form 
$\pi=\pi_0,\ldots,\pi_n$, by extending the set $\Prop$ 
to include one proposition $p_a$ per action $a$, and setting 
$\pi_i\doteq s_i\cup \{p_a\mid a = a_{i+1}\}$.
In this way, $\pi_i$ denotes the pair $(s_i, a_{i+1})$.
We will always assume this form, even if referring
to sequences of states and actions.
Given a finite trace $\pi$, an \LDLf formula $\varphi$, 
and a position $i$, 
we define when $\varphi$ \emph{is true} at step $i$,  
written $\pi,i\models\varphi$, by (mutual) induction, as follows:

\begin{itemize}\itemsep = 0pt
\item $\pi,i\models \ttrue$;
\item $\pi,i\models \lnot\varphi$ ~iff~ $\pi,i\not\models\varphi$;
\item $\pi,i\models \varphi_1\land\varphi_2$ ~iff~ $\pi,i\models\varphi_1$ and
  $\pi,i\models\varphi_2$;
\item $\pi,i\models \DIAM{\rho}\varphi$ ~iff~  there exists $i \le j \le\length(\pi)$
such that $\pi(i,j)\in \L(\rho)$ and $\pi,j\models\varphi$,
where the relation $\pi(i,j)\in\L(\rho)$ is as follows:
	\begin{itemize}\itemsep = 0pt
	\item $\pi(i,j)\in\L(\phi)$ if $j{=}i{+}1$, $j\le \length(\pi)$, and $\pi(i) \models \phi$\;\; ($\phi$ propositional);
	\item $\pi(i,j)\in\L(\varphi?)$ if $j=i\;\text{and}\;  \pi, i\models \varphi$;
	\item $\pi(i,j)\in\L(\rho_1+ \rho_2)$ if $\pi(i,j)\in\L(\rho_1)\;\text{or}\; \pi(i,j)\in\L(\rho_2)$;
	\item $\pi(i,j)\in\L(\rho_1; \rho_2)$ if  there exists $k$ such that $\pi(i,k)\in\L(\rho_1)$ and $\pi(k,j)\in\L(\rho_2)$;
	\item $\pi(i,j)\in\L(\rho^*)$ if $j=i$ or there exists $k$ such that
	  $\pi(i,k)\in\L(\rho)$ and $\pi(k,j)\in\L(\rho^*)$.
	\end{itemize}
\end{itemize}
Observe that if $i>\length(\pi)$, the above definitions still apply.
In particular,
$\BOX{\rho}\varphi$ is trivially true and $\DIAM{\rho}\varphi$ trivially false.

We say that a trace $\pi$ \emph{satisfies} an \LDLf formula $\varphi$,
written $\pi\models \varphi$, if $\pi,0\models \varphi$. Also,
sometimes we denote by $\L(\varphi)$ the set of traces that satisfy
$\varphi$: $\L(\varphi) = \{\pi\mid \pi \models \varphi\}$.

\LDLf is as expressive as \MSO over finite words. It captures \LTLf,
by seeing \emph{next} and \emph{until} as the abbreviations
$\Next \varphi \doteq \DIAM{\true} \varphi$ and
$\varphi_1 \Until \varphi_2 \doteq\DIAM{(\varphi_1?;\true)^*}
\varphi_2$, and any \REGEX $r$, with the formula $\DIAM{r}\Ended$,
where $\Ended\doteq \BOX{true?}\ffalse$ expresses that the trace has
ended.  Note that in addition to $\Ended$ we can also denote the last
element of the trace as $\Last\doteq \BOX{true}\Ended$  or equivalently
$\Last\doteq \BOX{true}\ffalse$. The latter has also an \LTLf-equivalent: $\lnot\Next\true$, instead $\Ended$ does not.

The properties mentioned at the beginning of the 
introduction can be expressed in \LDLf as follows:
$\BOX{\true^*} (\mathit{request}_p \limp \DIAM{\true^*}\mathit{coffee}_p)$
(all coffee requests from person $p$ will eventually be served);
$\BOX{\true^*}(\BOX{\mathit{openDoor}_d}\mathit{closeDoor}_d)$
(every time the robot opens door $d$ it closes it immediately after);
\[\begin{array}{l}
\DIAM{((\lnot \mathit{restrd}_a)^*;\mathit{permission}_a; (\lnot \mathit{restrd}_a)^*;\mathit{restrd}_a)^*; \\
\qquad(\lnot \mathit{restrd}_a)^*}\Ended
  \end{array}
\]
(before entering restricted area $a$ the robot must have permission for $a$).
While the \LTLf-equivalents of the first two formulas are immediate, i.e., 
$\Box(\mathit{request}_p \limp \Diamond\mathit{coffee}_p)$ and 
$\Box({openDoor}_d\limp \Next\mathit{closeDoor}_d)$, 
that for the third one is not.



We can associate each \LDLf formula $\varphi$ with an (exponentially large)
\NFA $A_\varphi$ that accepts exactly the traces satisfying $\varphi$.  
A simple direct algorithm (\algoname) for computing the \NFA
given  the \LDLf formula is reported below.
Its correctness relies on the fact that \myi 
every \LDLf formula $\varphi$ can be associated with a polynomial
\emph{alternating automaton on words} (\AFW)
$\A_\varphi$ accepting exactly the traces that satisfy $\varphi$
\cite{DegVa13}, and \myii every \AFW can be transformed into an
\NFA, see, e.g., \cite{DegVa13}. 
 
The algorithm assumes that the \LDLf formula
is in \emph{negation normal form} (NNF), 
i.e., with negation symbols occurring only in front of propositions
(any \LDLf formula can be rewritten in NNF in
linear time), and that 
$\Prop$ includes the special proposition
$\Last$, denoting the last element of the trace.
Let $\delta$ be the following auxiliary function, which takes in input 
an \LDLf formula $\psi$  (in NNF)
and a propositional interpretation $\Pi$ for $\Prop$ 
(including $\Last$), and returns a positive
boolean formula whose atoms are \emph{quoted} $\psi$ subformulas:
\allowdisplaybreaks
\[
\scriptsize
\begin{array}{rcl}
\delta(\ttrue,\Pi) & = &\true\\
\delta(\ffalse,\Pi) & = &\false\\
\delta(\phi,\Pi) &=& \left\{\hspace{-1ex}\begin{array}{l}
\true \mbox{ if } \Pi\models \phi \\
\false \mbox{ if } \Pi\not\models \phi
\end{array}\right.\quad \mbox{($\phi$ prop.)}\\
\delta(\varphi_1\land\varphi_2,\Pi) &=& 
\delta(\varphi_1,\Pi) \land \delta(\varphi_2,\Pi)\\
\delta(\varphi_1\lor\varphi_2,\Pi) &=& 
\delta(\varphi_1,\Pi) \lor \delta(\varphi_2,\Pi)\\
\delta(\DIAM{\phi}\varphi,\Pi) &=& 
\left\{\hspace{-1ex}\begin{array}{l}
\expand(\varphi) \mbox{ if } \Last \not \in \Pi \mbox{ and } \Pi \models \phi \quad \mbox{($\phi$  prop.)}\\
\delta(\expand(\varphi), \epsilon)  \mbox{ if } \Last \in \Pi \mbox{ and } \Pi \models \phi \\
\false \mbox{ if } \Pi\not\models \phi
\end{array}\right.\\[5pt]
\delta(\DIAM{\psi?}{\varphi},\Pi) &=& 
\delta(\psi,\Pi) \land \delta(\varphi,\Pi)\\
\delta(\DIAM{\rho_1+\rho_2}{\varphi},\Pi) &=& 
\delta(\DIAM{\rho_1}\varphi,\Pi)\lor\delta(\DIAM{\rho_2}\varphi,\Pi)\\
\delta(\DIAM{\rho_1;\rho_2}{\varphi},\Pi) &=& 
\delta(\DIAM{\rho_1}\DIAM{\rho_2}\varphi,\Pi)\\
\delta(\DIAM{\rho^*}\varphi,\Pi) &=& 
\delta(\varphi,\Pi) \lor 
\delta(\DIAM{\rho}\fff_{\DIAM{\rho^*}\varphi},\,\Pi)\\
\delta(\BOX{\phi}\varphi,\Pi) &=& 
\left\{\hspace{-1ex}\begin{array}{l}
   \varphi \mbox{ if } \Last \not \in \Pi \mbox{ and } \Pi \models \phi \quad \mbox{($\phi$ prop.)}\\
 \delta(\varphi,\epsilon) \mbox{ if } \Last \in \Pi \mbox{ and } \Pi \models \phi \\
    \true \mbox{ if } \Pi\not\models \phi
\end{array}\right.\\[5pt]
\delta(\BOX{\psi?}{\varphi},\Pi) &=& 
\delta(\nnf(\lnot\psi),\Pi) \lor \delta(\varphi,\Pi)\\
\delta(\BOX{\rho_1+\rho_2}{\varphi},\Pi) &=& 
\delta(\BOX{\rho_1}\varphi,\Pi)\land\delta(\BOX{\rho_2}\varphi,\Pi)\\
\delta(\BOX{\rho_1;\rho_2}{\varphi},\Pi) &=& 
\delta(\BOX{\rho_1}\BOX{\rho_2}\varphi,\Pi)\\
\delta(\BOX{\rho^*}\varphi,\Pi) &=& 
\delta(\varphi,\Pi) \land 
\delta(\BOX{\rho}\ttt_{\BOX{\rho^*}\varphi},\,\Pi)\\
\delta(\fff_{\psi},\Pi) & = & \false\\
\delta(\ttt_{\psi},\Pi) & = & \true
\end{array}
\]
where $\expand(\varphi)$ recursively replaces in $\varphi$ all occurrences of atoms of the form $\ttt_\psi$ and $\fff_\psi$ by $\expand(\psi)$; and $\delta(\ttrue,\epsilon)$ is defined as:
\[\scriptsize \begin{array}{rcl}
\delta(\ttrue,\epsilon) &=& \true\\
\delta(\ffalse,\epsilon) &=& \false\\
\delta(\phi,\epsilon) &=& \false  \quad \mbox{($\phi$  prop.)}\\
\delta(\DIAM{\rho\varphi},\epsilon) &=&\false\\
\delta(\BOX{\rho}\varphi,\epsilon) &= &\true
\end{array}
\begin{array}{rcl}
\delta(\varphi_1\land\varphi_2,\epsilon) &=& 
\delta(\varphi_1,\epsilon) \land \delta(\atomize{\varphi_2},\epsilon)\\
\delta(\varphi_1\lor\varphi_2,\epsilon) &=& 
\delta(\varphi_1,\epsilon) \lor \delta(\varphi_2,\epsilon)\\
\delta(\fff_{\varphi},\epsilon) & = &\false \quad \mbox{(for any $\varphi$)}\\
\delta(\ttt_{\varphi},\epsilon) & = &\true \quad \mbox{(for any $\varphi$)}
\end{array}
\]
\noindent 
The \NFA $\A_\varphi$ for an \LDLf formula $\varphi$ is then 
built in a forward fashion as shown below, where:
states of $\A_\varphi$ are sets of atoms (recall that each atom is
a quoted $\varphi$ subformula) to be interpreted as conjunctions;
the empty conjunction $\emptyset$ stands for $\true$;
$q'$ is a set of quoted subformulas of $\varphi$ denoting a
minimal interpretation such that 
$q'\models \bigwedge_{(\psi\in q)} \delta(\psi,\Pi)$
(notice that we trivially have
$(\emptyset,a,\emptyset)\in\varrho$ for every $a\in\Sigma$).

\algrenewcommand\algorithmicindent{1em}
\smallskip
\fbox{
\begin{minipage}{7.5cm}
  \begin{algorithmic}[1]
    \small 
\State\textbf{algorithm} \algoname() \\\textbf{input} \LDLf formula
    $\varphi$ \\\textbf{output} \NFA $\A_\varphi =
    (2^\Prop,\S,\{s_0\},\varrho,\{s_f\})$ 
 \State $s_0  \gets \{\varphi\}$
    \Comment{single initial state} 
  \State $s_f \gets \emptyset$ \Comment{single final state}  
  \State $\S \gets \{s_0,s_f\}$, $\varrho \gets \emptyset$ 
 \While{($\S$ or $\varrho$ change)} 
\If{($q\in \S$ and $q'\models
      \bigwedge_{(\psi\in q)} \delta(\psi,\Pi)$)}

    \State $\S \gets \S \cup \{q'\}$ \Comment{update set of states}
    \State $\varrho \gets \varrho \cup
   \{ (q,\Pi,q')\}$ \Comment{update transition relation}
    \EndIf 
    \EndWhile
 \end{algorithmic} 
\end{minipage}
}

\begin{theorem}{\cite{DegVa15}}
Algorithm  \algoname\ terminates in at most an exponential number of steps, 
and generates a set of states $\S$ whose size
is at most exponential in the size of the formula $\varphi$. 
\end{theorem}

Note that one can remove the proposition $\Last\in\Prop$ by
suitably adding an extra final state to $\A_\varphi$ \cite{DegVa15}.
The \NFA $\A_\varphi$ is correct, that is:

\begin{theorem}{\cite{DegVa15}}
For  every finite trace $\pi$:
$\pi\models\varphi \mbox{ iff } \pi\in L(\A_\varphi)$.
\end{theorem}

Finally, we can transform the \NFA $\A_\varphi$ into a \DFA 
in exponential time, following
the standard procedure, and then possibly 
put it in (the unique) minimal form, in polynomial time
\cite{RaSc59}. 
Thus,  we can transform any \LDLf formula into a
\DFA of double exponential size. While this is a
worst-case complexity, in most cases the size of the \DFA is actually
manageable~\cite{TaVa05}.



\section{Specifying Decision Processes with Non-Markovian Rewards}\label{sec:mdb-basic-partial}

In this section we extend MDPs with \LDLf-based reward functions resulting in 
a non-Markovian reward decision process (NMRDP). Then, we show how to
construct an equivalent extended MDP with Markovian rewards. 

A non-Markovian reward decision process (NMRDP) is a tuple $\M=\langle S,A,Tr,R\rangle$,
where $S,A$ and $Tr$ are as in an MDP, and $R$ is redefined as $R:(S\times A)^* \rightarrow \mathbb{R}$. The reward
is now a real-valued function over finite state-action sequences.
Given a (possibly infinite) trace $\pi$, the {\em value\/} of $\pi$ is:
$$v(\pi) =\sum_{i=1}^{|\pi|} \gamma^{i-1}R(\langle \pi(1),\pi(2),\ldots,\pi(i)\rangle),$$ where
$0<\gamma\leq 1$ is the discount factor and $\pi(i)$ denotes the pair $(s_{i-1},a_i)$. Since every policy $\rho: S^*\rightarrow A$ 
induces a distribution over the set of possible infinite traces, we can
now define the value of a policy $\rho$ given an initial state $s_0$ to be
$$v^{\rho}(s) = E_{\pi\sim\M,\rho,s_0} v(\pi)$$ That is, $v^{\rho}(s)$ is the expected value of infinite traces,
where traces are distributed according to the distribution over traces defined by the initial state $s_0$,
the transition function $Tr$, and the policy $\rho$. 

Specifying a non-Markovian reward function explicitly is cumbersome and unintuitive, even if we only
want to reward a finite number of traces. But, typically, we want to reward behaviors that
correspond to various patterns. \LDLf\ provides us with an intuitive and convenient language
for specifying $R$ implicitly, using a set of pairs $\{(\varphi_i,r_i)_{i=1}^{m}\}$. Intuitively, if the current
trace is $\pi=\langle s_0,a_1,\ldots,s_{n-1},a_n\rangle$, the agent receives at $s_n$ a reward  $r_i$ for every formula
$\varphi_i$ satisfied by $\pi$. 
Formally: $$R(\pi) = \sum_{1\leq i\leq m: \pi\models\varphi_i} r_i$$
From now on, we shall assume that $R$ is thus specified.

Coming back to our running example, we could have the following formulas $\varphi_i$,
each associated with reward $r_i$:\\
$\varphi_1 =  \BOX{\true^*} (\mathit{request}_p \limp \DIAM{\true^*}\mathit{coffee}_p)$\\
$\varphi_2 = \BOX{\true^*}(\BOX{\mathit{openDoor}_d}\mathit{closeDoor}_d)$\\
$\varphi_3 = 
\begin{array}{l}
\DIAM{((\lnot \mathit{restrd}_a)^*;\mathit{permission}_a; 
(\lnot \mathit{restrd}_a)^*;\mathit{restrd}_a)^*; \\
\qquad(\lnot \mathit{restrd}_a)^*}\Ended
\end{array}
$

\section{Building an Equivalent Markovian Model}
When the rewards are Markovian, one can compute $v^{\rho}$ (for stationary $\rho$) and
an optimal policy $\rho^*$ using Bellman's  dynamic programming equations~\cite{Puterman}.
However, this is not the case when the reward is non-Markovian and the policy is non-stationary.
The standard solution for this problem is to formulate an extended MDP in which the rewards are
Markovian, that is {\em equivalent} to the original NMRDP~\cite{BBG96,ThiebauxGSPK06}. 

\begin{definition}[\cite{BBG96}]
An 
NMRDP $\M=\langle S,A,Tr,R\rangle$ is equivalent to an extended MDP $\M'=\langle S',A,Tr',R'\rangle$
if there exist two functions $\tau:S'\rightarrow S$ and $\sigma:S\rightarrow S'$ such that
\begin{enumerate}
\item $\forall s\in S: \tau(\sigma(s)) = s$;
\item $\forall s_1,s_2\in S$ and $s_1'\in S'$: if $Tr(s_1,a,s_2)>0$ and
$\tau(s_1')=s_1$, there exists a unique $s_2'\in S'$  such that $\tau(s'_2) = s_2$
and $Tr(s'_1,a,s'_2) = Tr(s_1,a,s_2)$;
\item For any feasible trajectory $\langle s_0,a_1,\ldots,s_{n-1},a_n\rangle$ of $\M$ and
$\langle s'_0,a_1,\ldots,s'_{n-1},a_n\rangle$ of $\M'$, such that $\tau(s'_i)=s_i$ and
$\sigma(s_0)=s'_0$, we have $R(\langle s_0,a_1,\ldots,s_{n-1},a_n\rangle) = 
R'(\langle s'_0,a_1,\ldots,s'_{n-1},a_n\rangle)$.
\end{enumerate}
\end{definition}
As in previous work, we restrict our attention to extended MDPs such that $S'=Q\times S$,
for some set $Q$.

%
%
%
%
%
Given an NMRDP $\M = \langle S,A,Tr,R\rangle$,  we now show how to construct an equivalent extended MDP.
For each reward formula $\varphi_i$, we consider the corresponding (minimal) \DFA
$\A_{\varphi_i} = (2^{\Prop}, Q_i, q_{i0}, \delta_i,F_i)$,
where:
\begin{itemize}
\item $2^{\Prop}$ is the input alphabet of the automaton;
\item $Q_i$ is the finite set of states;
\item $q_{i0} \in Q_i$ is the initial state;
\item $\delta_i:Q\times 2^\Prop\rightarrow Q$ is the deterministic transition function
(which is total); 
\item $F_i\subseteq Q_i$ are the accepting
states.
\end{itemize}

\smallskip
We now define the equivalent extended MDP $\M' = \langle S',A',Tr',R'\rangle$
where:
\begin{itemize}
\item $S'=Q_1\times\cdots\times Q_m\times S$ is the set of states;
\item $A'=A$;
\item $\Tr' : S'\times A' \times S'\rightarrow [0,1]$ is defined as follows:
\[
\begin{array}{l}
\Tr'(q_1,\ldots,q_m, s, a, q'_1,\ldots,q'_m, s') = {}\\
\quad\left\{
\begin{array}{ll}
Tr(s,a,s') &\mbox{if } \forall i:\delta(q_i,s) = q'_i\\
0 & \mbox{otherwise}; 
\end{array}\right.
\end{array}
\] 
\item $R': S'\times A\rightarrow 
\mathbb{R}$ is defined as:
\[
R(q_1,\ldots,q_m, s, a) = 
\sum_{i: \delta(q_i,s) \in F_i} r_i
\] 
\end{itemize}
That is, the state space is a product of the states of the original MDP and the various automata. 
The actions set is the same.
Given action $a$, the $s$ component of the state progresses according to the original MDP dynamics, and
the other components progress according to the transition function of the corresponding automata.
Finally, in every state, and for every $1\leq i\leq m$, the agent receives the reward associated with $\varphi_i$ if the FSA $\A_{\varphi_i}$ reached a final state.


\begin{theorem}
The NMRDP $\M= \langle S,A,Tr,R\rangle$ is equivalent to the extended MDP $\M'= \langle S',A',Tr',R'\rangle$.
\end{theorem}
\proof Recall that every $s'\in S'$ has the form $(q_1,\ldots,q_m,s)$. 
Define $\tau(q_1,\ldots,q_m,s)=s$. Define $\sigma(s) = (q_{10},\ldots,q_{m0},s)$. We have $\tau(\sigma(s))=s$.
Condition 2 is easily verifiable by inspection. 
For condition 3, consider a possible trace $\pi=\langle s_0,a_1,\ldots,s_{n-1},a_n\rangle$.
We use $\sigma$ to obtain $s'_0=\sigma(s_0)$ and given $s_i$, we
define $s'_i$ (for $1\leq i < n$)  to be the unique state $(q_{1i},\ldots,q_{mi},s_i)$ such that
$q_{ji} = \delta_i(q_{ji-1},a_i)$ for all $1\leq j\leq m$. We now have a corresponding possible
trace of $\M'$, i.e., $\pi'=\langle s'_0,a_1,s'_1\ldots,s'_{n-1},a_n\rangle$. This is the only feasible trajectory of $\M'$ that
satisfies Condition  3.
The reward at $\pi=\langle s_0,a_1,s_1\ldots,s_{n-1},a_n\rangle$ depends only on whether or not each formula $\varphi_i$ is
satisfied by $\pi$. However, by construction of the automaton $\A_{\varphi_i}$ and the transition function $Tr'$,
$\pi\models\varphi_i$  {\em iff} $s'_{n-1}=(q_1,\ldots,q_m,s'_n)$ and $q_i\in F_i$.
\qed

Let $\rho'$ be a policy for the Markovian $\M'$. It is easy to define an equivalent policy on $\M$: Let $\pi=\langle s_0,a_1,s_1\ldots,s_{n-1},a_n\rangle$ be the current history of
the process. Let $q_{in}$ denote the current state of automaton $\A_{\varphi_i}$ given input $\pi$. 
Define $\rho(\pi) := \rho'(q_{1n},\ldots,q_{mn},s_n)$.

\begin{theorem}[\cite{BBG96}]
Given an NMRDP $\M$, let $\rho'$ be an optimal policy for an equivalent MDP $\M'$. Then,
policy $\rho$ for $\M$ that is equivalent to $\rho'$ is optimal for $\M$.
\end{theorem}

\section{Minimality and Compositionality}
One advantage of our construction is that it benefits from two types of minimality and from compositionality.
The Markovian model is obtained by taking the synchronous product of the original MDP and an FSA  
that is itself the synchronous product of smaller FSAs, one for each formula. 
We can apply the simple, standard automaton
minimization algorithm to obtain a minimal automaton, thus obtaining a minimal MDP. But even better, as we 
show below, it is enough to ensure that each FSA $\A_{\varphi_i}$ in the above construction is minimal to ensure the
overall minimality of the extended MDP.
\begin{theorem}
If every automaton $\A_{\varphi_i}$ ($1\leq i\leq m$) is minimal then the extended MDP defined
above is minimal.
\end{theorem}
\proof Let $\A_s$ be the synchronous product of $\A_{\varphi_i}$
($1\leq i\leq m$). We show that no two distinct states of the
synchronous product $\A_s$ are equivalent, and therefore, all of
them are needed, hence the thesis.

Suppose that there are two distinct states of the synchronous
product $\A_s$  that are equivalent. Then, being $\A_s$ a \DFA, such two states are
bisimilar. 
Two states of $\A_s$  are bisimilar (denoted by $\sim$) iff:
$(q_1,\ldots,q_n)  \sim (t_1,\ldots,t_m)$  implies
\begin{itemize}
\item for all $i$.   $q_i \in F_i$ iff $t_i \in F_i$;

\item for all $a$.  $\delta_s(q_1,\ldots,q_m) = (q'_1,\ldots,q'_m)$
  implies $\delta_s(t_1,\ldots,t_m,a) = (t'_1,\ldots, t'_m)$ and
  $(q'_1,\ldots,q'_m)\sim (t'_1,\ldots,t'_m)$;

\item for all $a$.  $\delta_s(t_1,\ldots,t_m) = (t'_1,\ldots,t'_m)$
  implies $\delta_s(q_1,\ldots,q_m,a) = (q'_1,\ldots, q'_m)$ and
  $(q'_1,\ldots,q'_m)\sim (t'_1,\ldots,t'_m)$.
\end{itemize}
             

Now we show that $(q_1,\ldots,q_m) \sim (t_1,\ldots,t_m)$ implies
$q_i = t_i$, for all $i$. To check this we show that the relation
``project on $i$'', $\Pi_i((q_1,\ldots, q_m) \sim (t_1,\ldots, t_m))$
extracting the $i$-th component on the left and on the right of $\sim$
is a bisimulation for states in $\A_i$. Indeed it is immediate to verify that
$\Pi_i((q_1,\ldots, q_m) \sim (t_1,\ldots,t_m))$  implies

\begin{itemize}
\item $q_i \in F_i$ iff $t_i \in F_i$;

\item for all $a$,
     $\delta_i(q_i ,a) =q'_i$ implies  $\delta_i(t_i,a) =t'_i$ and
               $\Pi_i((q'_1,\ldots, q'_m) \sim (t'_1,\ldots,t'_m))$;

\item for all $a$,
     $\delta_i(t_i ,a) =t'_i$ implies  $\delta_i(q_i,a) =q_i$ and
               $\Pi_i((q'_1,\ldots, q'_m) \sim (t'_1,\ldots,t'_m))$. 
\end{itemize}

Hence if there are two distinct states
$(q_1,\ldots,q_m) \sim (t_1,\ldots,t_m)$ then at least for one $i$ it
must be the case that $q_i$ and $t_i$ are distinct and bisimilar and hence
equivalent. But this is impossible since each \DFA $\A_{\varphi_i}$ is
minimal. 
\qed\\

In general, if we take the synchronous product of two minimised \DFA's we may be able to minimize it further. But since in our case we need to keep the final states of the different \DFA's distinct (to assign the proper rewards), as the proof of theorem above shows, 
no further minimization is possible. 



Observe that the above theorem also implies that the construction is compositional, and hence, incremental -- if we care for a new formula, we do not need
to change the MDP, but simply extend it with one additional component. 
If the original MDP was minimal and the new component is minimal, then so is the resulting MDP.

\cite{ThiebauxGSPK06} consider a different minimality criterion, {\em
  blind minimality}, which essentially says that,
given an initial state $s_0$ for our NMRDP $\M$, one can construct the
set of states reachable from $s_0$ in $\M'$ without having to generate
any unreachable extended state. In particular, this implies that one does
not generate the entire automaton for each formula, but construct only
its reachable states. Moreover, one can even focus on a subset
of reachable states that correspond to trajectories of interest.

We enjoy both notions. We can progress the extended MDP, starting
from the initial state, building it on the fly. But we can also start by
generating the reachable states of each automaton separately; minimize each
automaton, and take their synchronous product. While the automata are theoretically large
(as is the reachable state space), in practice, experience shows them to be
quite small. Once we have the minimal structure of the automaton,
we can progress the extended MDP working with the product of the MDP and automaton states.

\section{Getting rewards for complete traces only} 
We may want to reward an agent for its entire behavior rather than for each prefix of it.
This means that the value of a sequence $\pi=\langle s_0,a_1,s_1\ldots,s_{n-1},a_n\rangle$ is defined as follows:
\[ v(\pi)= \sum_{i:\pi\models\varphi_i} r_i 
\] 

Behaviors optimal w.r.t.~this definition will differ from ones that are optimal w.r.t.~the original definition in which rewards are collected
following each action. The point is that an agent must now attempt to make as many formulas true at once, as it does not get any ``credit'' for having
achieved them in the past.

Given an NMRDP $\M$ with the above reward semantics, we can easily generate an equivalent MDP
using the above construction, preceded by the following steps:
\begin{enumerate}
\item Add a special action {\em stop} to $A$.
\item Add a new proposition {\em done} to $S$.
\item No action is applicable in a state in which {\em done} is {\em true\/}.
\item The only effect of the {\em stop} action is to make {\em done}  be {\em true\/}.
\item Convert every reward formula $\varphi_i$ to $done\wedge\varphi_i$.
\end{enumerate}

%
%
%

Interestingly, when focussing on complete traces, our framework becomes an extension of Goal MDP planning that handles temporally extended goals, see, e.g, Chapter~6 and Chapter~4 of \cite{GeBo13}.

\section{Comparison with previous proposals}
\paragraph{Capturing \PLTL rewards.}

The setting proposed can be seen as an extension of~\cite{BBG96}. There, rewards are assigned to partial traces whenever the last state of the trace satisfies a past-\LTL (\PLTL) formulas. Without introducing explicitly \PLTL, but given a partial  trace  $\pi_0,\ldots,\pi_n$ we reverse it into $\pi_n,\ldots, \pi_0$ and evaluate it over the \LTLf formula $\varphi$ obtained from the \PLTL formula by simply replacing the past operators with the corresponding future operators (e.g., replace {\em since} with 
{\em eventually\/}). Then, the setting remains analogous to the one shown above.

In particular, we can construct the \NFA $A_\varphi$ associated with $\varphi$ and, 
instead of reversing the partial traces, reverse $A_\varphi$,  thus getting an \NFA $A^-_\varphi$, 
by simply reversing the edge directions and switching initial and final states. 
This can be done in linear time. If we now determinize (and minimize) $A^-_\varphi$,  getting the (minimal) \DFA $A^-_\varphi$, we can proceed exactly as above.  

Given the above essential equivalence of \PLTL and \LTLf, and the fact that \LDLf\ is strictly more expressive than \LTLf, we
conclude that our setting is \emph{strictly more} expressive than the one in \cite{BBG96}.
In principle, we could simply 
replace \PLTL with past-\LDLf. But defining properties in past-\LDLf is likely to be unnatural, since we would have to reverse the regular expressions in the eventualities, and since these have a procedural flavour, it would be somehow like reversing a program. 
In addition, since the automata construction algorithm is based on progression,  unlike~\cite{BBG96},
we can use information about the initial state to prevent the generation of unreachable states.

Finally, we note that in~\cite{BBG97}, the authors extend their work to handle NMRDPs in factored form and attempt to ensure that the 
extended state retains this factored form.
We note that our construction retains the original form of the MDP, whether factored or not, and generates a natural factored
extended state, using one factor per reward formula. 

\vspace{1ex}
\noindent {\bf Comparing with  \DFLTL rewards.} 
In \cite{ThiebauxKS02,GrettonPT03,ThiebauxGSPK06} a sophisticated temporal logic, called \DFLTL is introduced, which is able to specify explicitly when a partial (finite) trace gets a rewards.  The exact expressive power of \DFLTL has not been  assessed yet, and it is open whether it is able to capture \PLTL rewards of \cite{BBG96} and vice-versa.
%
As a result, it remains open to compare our setting, based on \LDLf, with \DFLTL. 
However, as \DFLTL is based on \LTL, which cannot capture \MSO, 
it would be rather surprising if  \DFLTL  was able to capture the \LDLf rewards proposed here. 

We can show, though, how some \DFLTL formulas can be expressed in 
\LDLf\footnote{For further examples see Appendix \ref{appendix:examples}.}. 
We consider the examples of~\cite{ThiebauxGSPK06}.
Some of these show how the \PLTL formulas used in~\cite{BBG96} can be expressed 
in \DFLTL. By the relationship between \PLTL and \LDLf discussed above, 
it is immediate that these admit an \LDLf equivalent formula. For instance,
$\lnot p\Until (p\land \$)$, which rewards only the first time $p$ is achieved,
is equivalent to the \PLTL formula $p\land \lnot \prev\past p$, which 
can be, in turn, rewritten in \LDLf as $\DIAM{\lnot p^*;p}\Ended$ or in \LTLf syntax as $\lnot p\Until (p\land \Last)$. 
There are also \DFLTL
formulas for which no equivalent  in \PLTL is reported. 
This is the case, e.g., of $\lnot q\Until ((\lnot p \land \lnot q)\lor(q\land \$))$,
which rewards the holding of $p$ until the occurrence of $q$, 
and whose \LDLf translation is $\DIAM{q^*;p}\Ended$  or, in \LTLf, $p\Until (q\land\Last)$.
Observe how simpler it is to have an intuition of the semantics
when using the \LDLf version compared with the \DFLTL one. In 
particular, the latter requires a rigorous application of the
semantics even to simply check that the property is
as claimed above.
In addition, various properties compactly expressible in \LDLf require much more sophisticated encoding when only standard temporal operators are used. 
Finally, computationally, we can offer the benefits of true minimality and
blind minimality (reachability) as well as compositionality.


\section{Conclusion}\label{sec:conclusion}

We presented a new language for specifying non-Markovian rewards in MDPs.
Our language is more expressive than previous proposals and being based on
a standard temporal logics of the future, is likely to be more intuitive to use.
We showed how to construct a minimal equivalent MDP, and since we rely
on general methods for tracking temporal formulas, the construction is
cleaner. Being based on progression, it can use information about the initial state to 
prune unreachable states.

One problem with non-Markovian rewards is that the reward is
only obtained when the entire sequence satisfies the property. This is
especially true if we wish to give rewards for complete traces only.
In that case, the reward comes as a "surprise" when the last action
$\Stop$ is chosen.  We can help solution algorithms if we can start
rewarding such behaviors even before the formula is satisfied, helping
to guide both search and learning algorithms towards better behaviors.
In future work we intend to examine the use of monitoring notions developed
for \LTLf\ and \LDLf~\cite{Bauer2010:LTL,DDGMM-BPM14,MMW11}. 
Using such monitors one
could extract early rewards that guide the process to get 
full rewards later. 

Another important direction for future work is exploiting non-Markovian rewards in reinforcement learning (RL)
to provide better guidance to the learning agent, as well as extending inverse RL methods to learn 
to assign non-Markovian rewards in a state. We are currently exploring this latter issue when
the set of formulas $\varphi_i$ is given, but the associated reward $r_i$ is unknown.

%
%
%


\clearpage

\bibliographystyle{named}
\bibliography{main-bib}

\clearpage
\appendix
\section{Appendix: Examples of Translations of \PLTL and \DFLTL formulas to \LDLf}\label{appendix:examples}
\section*{Examples of Translations of \PLTL and \DFLTL formulas to \LDLf} 

In this section, we show how formulas used in~\cite{BBG96} and in~\cite{ThiebauxGSPK06} can be 
encoded in \LTLf and \LDLf:

\begin{enumerate}
	\item A reward is offered only at the first state where a goal $G$ holds.
		\begin{itemize}
			\item	\PLTL: $G\land\lnot\prev\gpast G$
			\item \DFLTL: $\lnot G\Until (G\land\$)$
			\item \LTLf: $\lnot G\Until (G\land\Last)$
			\item \LDLf: $\DIAM{\lnot G^*; G}end$
		\end{itemize}
		
	\item A reward is offered at every state that follows $G$ (included):
		\begin{itemize}
			\item	\PLTL: $\past G$
			\item \DFLTL: $\Box (G\rightarrow \Box\$)$
			\item \LTLf: $\Diamond G$
			\item \LDLf: $\DIAM{true^*; G; true^*}end$
		\end{itemize}
		
	\item Achievement of $G$ is rewarded periodically, at most once every $k$ steps:
		\begin{itemize}
			\item	\PLTL: $G\land\lnot(\prev^{\leq k}G)$, where: 
				\begin{itemize}
					\item $\prev^{\leq k}G\doteq \bigvee_{i=1}^k\prev^i G$
					and $\prev^i G\doteq \underbrace{\prev\ldots\prev}_{i\text{ times}} G$
				\end{itemize}
			\item \DFLTL: $\Box((\Next^{k+1}G\land \Box_k\lnot G)\rightarrow\Next^{k+1}\$)$,
				where: 
				\begin{itemize}
					\item $\Next^i G\doteq \underbrace{\Next\ldots\Next}_{i\text{ times}} G$
					and $\Box_i G = \bigwedge_{j=1}^i
						\underbrace{\Next\ldots\Next}_{j\text{ times}} G$
				\end{itemize}
			\item \LTLf: $\Diamond(\Next^k(G\land\Last)) \land (\bigwedge_{j=0}^{k-1} \Next^j\lnot G)$
			\item \LDLf: $\DIAM{\lnot G^*;G;(\lnot G^k;\lnot G^*;G)^*)}end$,
				where: 
				\begin{itemize}
					\item $G^k=(\underbrace{G;\ldots;G}_{k\text{ times}})$
				\end{itemize}
			\end{itemize}

	\item Achievement of $G$ is rewarded whenever it occurs within $k$ steps 
	of a state where $\lnot G$ holds (we assume $k\geq 1$):
		\begin{itemize}
			\item	\PLTL: $G\land\prev^{\leq k}\lnot G$
			\item \DFLTL: $\Box(\lnot G\rightarrow \Box_k(G\rightarrow\$))$
			\item \LTLf: $\Diamond(\lnot G\land \Box_k(\Last\rightarrow G))$
			\item \LDLf: $\DIAM{true^*;\lnot G;G+((\lnot G^1+\ldots+\lnot G^k);G)}end$
		\end{itemize}
		
	\item A reward is issued whenever $G$ is achieved and followed immediately by $H$ and then 
	by $I$:
		\begin{itemize}
			\item	\PLTL: $\prev^2 G\land \prev H\land I$
			\item \DFLTL: $\Box((G\land \Next H\land \Next^2 I)\rightarrow\Next^2 \$)$
			\item \LTLf: $\Diamond(G\land\Next H\land \Next^2(I\land\Last))$
			\item \LDLf: $\DIAM{true^*;G;H;I}end$
		\end{itemize}
		
	\item Achievement of $G$ is rewarded whenever it follows $C$:
		\begin{itemize}
			\item	\PLTL: $G\land \prev C$
			\item \DFLTL: $\Box(C\rightarrow {\Next}{\Box}(G\rightarrow\$))$
			\item \LTLf: $\Diamond (C\land\Next\Diamond(G\land\Last))$
			\item \LDLf: $\DIAM{true^*;C;true^*;G}end$
		\end{itemize}
		
	\item Only the first achievement of $G$ that follows $C$ is rewarded:
		\begin{itemize}
			\item	\PLTL: $G\land \prev (\lnot G\Since C)$
			\item \DFLTL: $\Box(C\rightarrow \Next(\lnot G\Until(G\land\$)))$
			\item \LTLf: $\Diamond (C\land\lnot G\Until(G\land \Last))$
			\item \LDLf: $\DIAM{true^*;C;\lnot G;\lnot G^*;G}end$
		\end{itemize}

	\item $G$ is rewarded whenever it follows $C$ immediately:
		\begin{itemize}
			\item	\PLTL: $G\land \prev C$
			\item \DFLTL: $\Box((C\land \Next G) \rightarrow \Next \$)$
			\item \LTLf: $\Diamond (C\land\Next(G\land\Last))$
			\item \LDLf: $\DIAM{true^*;C;G}end$
		\end{itemize}
		
	\item Achievement of $G$ is rewarded whenever occurring within $k$ steps ($k\geq 1$) of $C$:
		\begin{itemize}
			\item	\PLTL: $G\land \prev^{\leq k} C$
			\item \DFLTL: $\Box(C\rightarrow \Box_k(G\rightarrow \$))$
			\item \LTLf: $\Diamond (G\land\Last\land(\bigvee_{i=0}^k\Next^k C))$
			\item \LDLf: $\DIAM{true^*;C;G+((\true^1+\ldots+\true^k);G)}end$
		\end{itemize}
		
	\item Only the first achievement of $G$ occurring within $k$ steps ($k\geq 1$) of $C$ is rewarded:
		\begin{itemize}
			\item	\PLTL: $G\land \prev^{\leq k} C\land (\lnot G\Since C)$
			\item \DFLTL: $\Box(C \land (\Next G \rightarrow\Next \$)\land (\Next \lnot G \land \Next^2 G\rightarrow\Next^2 \$)\land\ldots\land(\Next \lnot G \land\ldots\land \Next^k \lnot G\land 
			\Next^{k+1} G\rightarrow\Next^{k+1} \$))$
			\item \LTLf: $\Diamond(C\land \Box_k(\Last\leftrightarrow G))$
			\item \LDLf: $\DIAM{true^*;C;G+((\lnot G^1+\ldots+\lnot G^k);G)}end$
		\end{itemize}
		
	\item Reward is issued if $G$ has always been true:
		\begin{itemize}
			\item	\PLTL: $\gpast G$
			\item \DFLTL: $\$\Until \lnot Q$
			\item \LTLf: $\Box G$
			\item \LDLf: $\DIAM{G^*}end$
		\end{itemize}
		
	\item The holding of $C$ until $G$ is rewarded:
		\begin{itemize}
			\item	\PLTL: $G\land{\prev\gpast}C$
			\item \DFLTL: $\lnot G\Until((\lnot C\land\lnot G)\lor(G\land\$))$
			\item \LTLf: $C\Until(G\land\Last)$
			\item \LDLf: $\DIAM{C^*;G}end$
		\end{itemize}
\end{enumerate}

\section*{Additional Examples}
Here we present some examples of \LDLf formulas for which there seem not to be
any \LTLf translation.

\paragraph{} It is known that the \LDLf formula $\DIAM{(\ttrue\ttrue)^*}end$ (parity) is not expressible in \LTLf. 
Also the formula $\phi=\DIAM{(pr)^*}end$ seems to be so. Indeed, a reasonable 
(and perhaps the most natural) \LTLf candidate
for this formula would be a formula like the following:
$$\psi=\Last\lor \big(p \land \Box(p\rightarrow r) \land\Box(r \rightarrow (p\lor \Last))\big).$$

However, this formula does not capture $\phi$. To see this, consider the 
following trace:
$$\pi=\{p,r\}\{p,r\}\{p,r\}\{p,r\}.$$

While this satisfies $\phi$, it does not satisfy $\psi$, as the last occurrence of $p$
is not followed by $r$, as required by $\psi$.

We do not prove that the desired \LTLf formula does not exist but
we observe that this is very unlikely. Indeed, on traces of the 
same form as $\pi$ (with variable length), 
there is no way to distinguish the states on a 
local basis. Thus, the only way to check whether the property 
captured by $\phi$ is enforced seems to be checking whether a state is in even 
or odd position (states at odd position are required to 
satisfy $p$ and states at even position must satisfy $r$). 
But this cannot be the case, as  
\LTLf cannot express parity.

\paragraph{} Consider the example of the introduction, where a 
robot is rewarded if it accesses a restricted area only after having acquired the right
permission.
This can be captured in \LDLf by the following formula: 
$$\phi=\DIAM{(\lnot r^*;p;\lnot r^*; r)^*}{\Box}\lnot r,$$
where proposition $p$ stands for ``permission granted'' and $r$ for 
``restricted area entered''. This formula essentially says that 
any occurrence of $r$ must be preceded by at least one occurrence
of $p$ that is not already followed by another occurrence of $r$.
In other words, all occurrences of $p$ are `` consumed'' when $r$ is 
seen. Again, it is very unlikely that some \LTLf formula exists for this 
property, as \LTLf is not able to distinguish, in general, 
distinct occurrences of a same proposition along a run.

For an attempt to find an \LTLf formula, 
consider the following \LTLf formula:
$$\psi_4=(\Diamond r)\rightarrow \lnot(\lnot p\Until r),$$
which expresses that if an $r$ is seen in the future, it must be preceded by 
a $p$. This formula imposes a constraint on the first occurrence of $r$ only

We could then try with $\phi_5 = \Box\phi_4$, but this would not work 
either. Indeed, $\phi_5$ is not satisfied even by the following simple trace $\tau = \{p\}\{\}\{r\}$. In this case, indeed, $\phi_4$ would not be satisfied in the second state.
Formula $\phi_5$ shows a problem that occurs often when trying to 
capture an \LDLf formula in \LTLf, i.e., the impossibility of distinguishing 
the different occurrences of a same property.

\section*{Examples of Procedural Constraints in \LDLf} 

Interesting examples of \LDLf formulas, can be obtained by considering that \LDLf, differently from \LTLf, is able to easily express procedural
constraints \cite{DegVa15,FritzM07,BaierFBM08}.
In particular, we can introduce a sort of propositional variant of \textsc{Golog}
\cite{LRLLS97}:
\[\begin{array}{lcl}
\delta &::=& A \mid \varphi? \mid  \delta_1 + \delta_2 \mid \delta_1; \delta_2 \mid \delta^*\mid {}\\
&& \textbf{if}\ \phi\ \textbf{then}\ \delta_1\ \textbf{else}\ \delta_2 \mid \textbf{while}\ \phi\ \textbf{do}\ \delta
\end{array}
\]
Note that \textbf{if} and \textbf{while} can be seen as abbreviations for \LDLf path expression \cite{FiLa79}, namely:
\[\begin{array}{rcl}
\textbf{if}\  \phi\ \textbf{then}\ \delta_1 \textbf{else}\ \delta_2
& \doteq & (\phi?;\delta_1) + (\lnot\phi?;\delta_2)\\
 \textbf{while}\  \phi\ \textbf{do}\ \delta & \doteq &\ (\phi?;\delta)^*;\lnot\phi?
\end{array}
\]

Hence, we can assign rewards to  \LDLf formulas expressing the traces satisfying procedural constrains. For example:
\[\begin{array}{l}
\BOX{\true^*}\DIAM{\textbf{while} (\textit{cold}  \land \textit{heatingOn}))\ \textbf{do} \\
\qquad\qquad \lnot \textit{turnOffHeating}^*; \textit{heat}}\Ended
\end{array}
\]
which says that at every point, while it is cold and the heating is on then heat, possibly allowing other action except turning off heating; and 

\[\begin{array}{l}
\DIAM{\textbf{while} (\true) \textbf{do} \\
\qquad \textbf{if}\ (\textit{cold}  \land \textit{windowOpen})) \textbf{then}\\
\qquad \qquad\textit{closeWindow};\\
\qquad \qquad\textit{turnOnFirePlace} +  \textit{turnOnHeating}}\Ended\\
  \end{array}
  \]
which says that all along if it is cold and the window is open, then immediately close the window and either turn on the fire place or the heating system (differently form before, no other actions can interleave this sequence).

\end{document}